\documentclass[a4paper, 10pt, conference]{ieeeconf}      
\usepackage{FG2023}
\usepackage{soul}

\FGfinalcopy 

\IEEEoverridecommandlockouts                              
\def\FGPaperID{****} 

\title{\LARGE \bf
Pain Detection in Masked Faces during Procedural Sedation
}

\author{\parbox{16cm}{\centering
    {\large Y. Zarghami$^{1, 2}$, S. Mafeld$^{3}$, A. Conway$^{4, 5}$, and B. Taati$^{1, 2}$}\\
    {\normalsize
    $^1$ KITE Research Institute, Toronto Rehabilitation Institute - University Health Network, Toronto, ON, Canada\\
    $^2$ Department of Computer Science - University of Toronto, Toronto, ON, Canada \\
    $^3$ Division of Vascular and Interventional Radiology, Department of Medical Imaging, University of Toronto, Toronto, ON, Canada \\
    $^4$ Lawrence S. Bloomberg Faculty of Nursing, University of Toronto, Toronto, ON, Canada \\
    $^5$ Peter Munk Cardiac Centre, University Health Network, Toronto, ON, Canada}}
}

\usepackage[utf8]{inputenc} 
\usepackage[T1]{fontenc}    
\usepackage{hyperref}       
\usepackage{url}            
\usepackage{booktabs}       
\usepackage{amsfonts}       
\usepackage{nicefrac}       
\usepackage{microtype}      
\usepackage{xcolor}         
\usepackage{amsmath}
\usepackage{graphicx, caption}
\usepackage{float}
\usepackage{subfig}   

\usepackage{fancyhdr}

\fancypagestyle{plain}{%
  \fancyhf{}
  \fancyfoot[C]{\iffloatpage{}{\thepage}}
  }
\begin{document}

\ifFGfinal
\thispagestyle{empty}
\pagestyle{empty}
\else
\author{Anonymous FG2023 submission\\ Paper ID \FGPaperID \\}
\pagestyle{plain}
\fi
\maketitle

\begin{abstract}

Pain monitoring is essential to the quality of care for patients undergoing a medical procedure with sedation. An automated mechanism for detecting pain could improve sedation dose titration. Previous studies on facial pain detection have shown the viability of computer vision methods in detecting pain in unoccluded faces. However, the faces of patients undergoing procedures are often partially occluded by medical devices and face masks. A previous preliminary study on pain detection on artificially occluded faces has shown a feasible approach to detect pain from a narrow band around the eyes. This study has collected video data from masked faces of 14 patients undergoing procedures in an interventional radiology department and has trained a deep learning model using this dataset. The model was able to detect expressions of pain accurately and, after causal temporal smoothing, achieved an average precision (AP) of 0.72 and an area under the receiver operating characteristic curve (AUC) of 0.82. These results outperform baseline models and show viability of computer vision approaches for pain detection of masked faces during procedural sedation. Cross-dataset performance is also examined when a model is trained on a publicly available dataset and tested on the sedation videos. The ways in which pain expressions differ in the two datasets are qualitatively examined.

\end{abstract}

\section{INTRODUCTION}

\subsection{Motivation}

Pain monitoring is essential to the quality of patient care~\cite{painQualityofCare} as unnoticed and untreated pain can have severe physical and psychological risks~\cite{dementiaPaper}. Certain populations -- such as neonates, people with dementia, and those under sedation -- cannot provide sufficient or unambiguous self-report of pain due to cognitive limitations that reduce their ability to verbally communicate~\cite{neonatePain, dementiaPain,sedationPain, neonatePain2}. Clinically valid methods of pain assessment have been developed to alleviate this problem~\cite{facs, PACSLAC-II}; however, due to a shortage of staff equipped with pain-assessment skills, an automated method of pain detection is desirable. Partial occlusion of patients' faces, due to the requirement for patients to wear masks in certain care units, is an added complication to the automatic pain-monitoring process. This paper aims to address the problem of pain detection in patients undergoing a medical procedure with sedation, who have their faces partially occluded by medical devices, such as oxygen delivery devices (nasal cannulae and masks), or face masks (as has been required during the COVID-19 pandemic). Such a method can be used in improving sedation dose titration in these patients and can enhance the overall quality of care.

\subsection{Clinically Valid Assessment of Pain}

The Prkachin and Solomon Pain Index (PSPI) is a clinically valid method of coding facial pain assessment~\cite{pspi}. The PSPI score is built upon  the Facial Action Coding System (FACS), which itself ascribes action units to different facial muscle movements~\cite{facs}. The PSPI uses a subset of facial action units to calculate a score in the 0-16 range, where 0 and 16 correspond with no and maximum levels of expressed pain, respectively.

\subsection{Related Work}

A significant amount of work has been done on automated pain expression detection, the majority of which use the publicly available UNBC-McMaster Shoulder Pain Expression Archive~\cite{unbc} and/or the BioVid Heat Pain Database~\cite{biovid}. Computer vision and machine learning techniques have also been employed for pain-recognition in specific populations, e.g. 
in neonates~\cite{neonatePaper} and 
in people with dementia~\cite{dementiaPaper}.

\begin{figure}
\centering
\captionsetup{width=1\linewidth}
    \includegraphics[width=.5\linewidth]{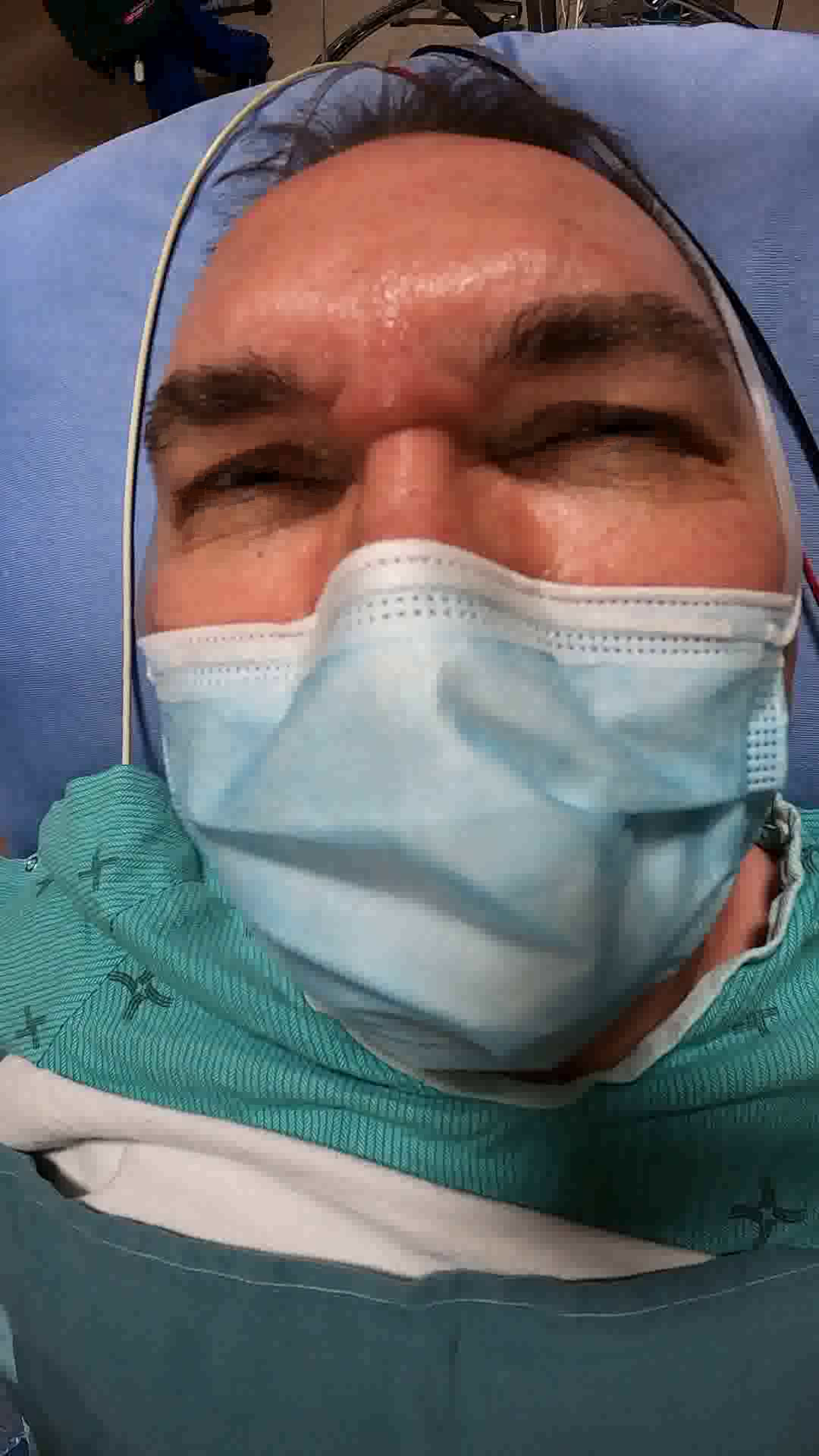}
    \caption{Sample pain frame from the sedation dataset.}
    \label{fig:SampleFullImage}
\end{figure}

Of particular relevance to this study, Ashraf et al. conducted a proof-of-concept analysis on pain-recognition in partially occluded faces~\cite{occludedFacesPaper}. To simulate partial occlusion, they extracted a narrow band around the eyes of the image frames in the UNBC-McMaster dataset. Three feature representation methods -- discrete cosine transform (DCT), local binary pattern (LBP), and histogram of oriented gradients (HOG) -- were used on the eye region frames and a linear support vector machine (SVM) binary classifier was trained on the extracted features. Results showed the feasibility of  detecting pain on partially occluded faces, albeit with lower accuracy than when analyzing unoccluded faces.

\subsection{Contributions}

Our work investigates the feasibility of using computer vision models for the automatic detection of pain in masked faces of patients undergoing procedural sedation. Whereas Ashraf et al.'s preliminary analysis~\cite{occludedFacesPaper} previously examined this in simulated data, our work is the first to use a video dataset of patients from the target population, undergoing a medical procedure with sedation in a hospital setting with real face occlusions. We fine-tune a pretrained deep learning model on this dataset and, in leave-one-person-out cross-validation, show that this model outperforms Ashraf et al.'s baselines trained on the same data. 
We investigate whether adding variability to our training set using cropped faces from the publicly available portion of the UNBC-McMaster dataset improves its performance. We also report cross-dataset performance of a model trained solely on the UNBC-McMaster dataset cropped to simulate partial occlusion. Finally, we qualitatively examine the ways pain expressions differ in the two datasets.

\section{DATA}

\subsection{Datasets Used}

We used two datasets to conduct our experiments. 
The primary dataset was drawn from a prospective observational study conducted in the Interventional Radiology department at a large academic hospital in North America. Patients undergoing elective procedures without general anesthesia from April to September 2021 were recruited. All participants provided written informed consent and the study was approved by the University Health Network Research Ethics Board (20-5900). 
Participants' faces were recorded using a GoPro camera with a flexible attachment during their procedure.  All participants' mouth and lower face regions were occluded by a face mask or a medical device, during the entire procedure. Videos were recorded with 1920$\times$1080 resolution, and at either 30 frames per seconds (fps), for 7 participants, or at 60 fps, for the remaining 7 participants.
Concurrently, a Research Nurse rated the adequacy of sedation in real-time using a sedation assessment scale~\cite{cravero2017validation}, which incorporated an assessment of pain. Binary 'pain' or 'no pain' ratings, derived from the nurse's sedation state assessments, were used as 'ground truth' for this analysis. Although attempts were made during procedures to optimize the view of facial expressions, there were periods of time where procedure positioning requirements obstructed the camera. As such, all video recordings were visually inspected to exclude segments in which participant's face was not in view. Figure~\ref{fig:SampleFullImage} illustrates a sample (pain) frame from this dataset.
A total of 14 participants were recruited for this study. A break-down of the number of image frames for each participant can be found in the rightmost column of Table~\ref{table:participantSmoothedResults}.

The UNBC-McMaster Shoulder Pain Expression Archive was also used to augment our primary dataset and  increase the size of our training set. The publicly available portion of this dataset includes videos of 25 participants with shoulder pain that undergo range-of-motion assessments on their affected and unaffected shoulders for the duration of the video. The videos are recorded at 30 fps and in Quarter VGA (240$\times×$320) resolution. The dataset contains 48,391 image frames in total, \emph{i.e.} approximately one tenth of the primary dataset in size, with 1936 ± 837 image frames per participant. This dataset has frame-by-frame annotations of relevant FACS codes, which are used to calculate and label each frame by its PSPI score. 

\subsection{Data Preprocessing}

The same preprocessing steps were applied to image frames from both the UNBC-McMaster and sedation datasets. The videos, particularly those from the sedation dataset, include motions of the head and the body and shifts in the background. To reduce these variabilities, the image frames were cropped around the unoccluded regions of the face. To create the crops for each individual frame from the UNBC-McMaster dataset, 2D facial landmark estimates from  the Face Alignment model (FAN)~\cite{2DFAN} were used. This facial landmark detection model did not generate reliable results in the sedation dataset with masked faces. The MediaPipe face mesh yielded reliable results and was used instead on the sedation dataset to create the crops.  

\begin{figure}
\centering
\captionsetup{width=1\linewidth}
    \begin{minipage}{5.2cm}
    \includegraphics[width=.9\textwidth]{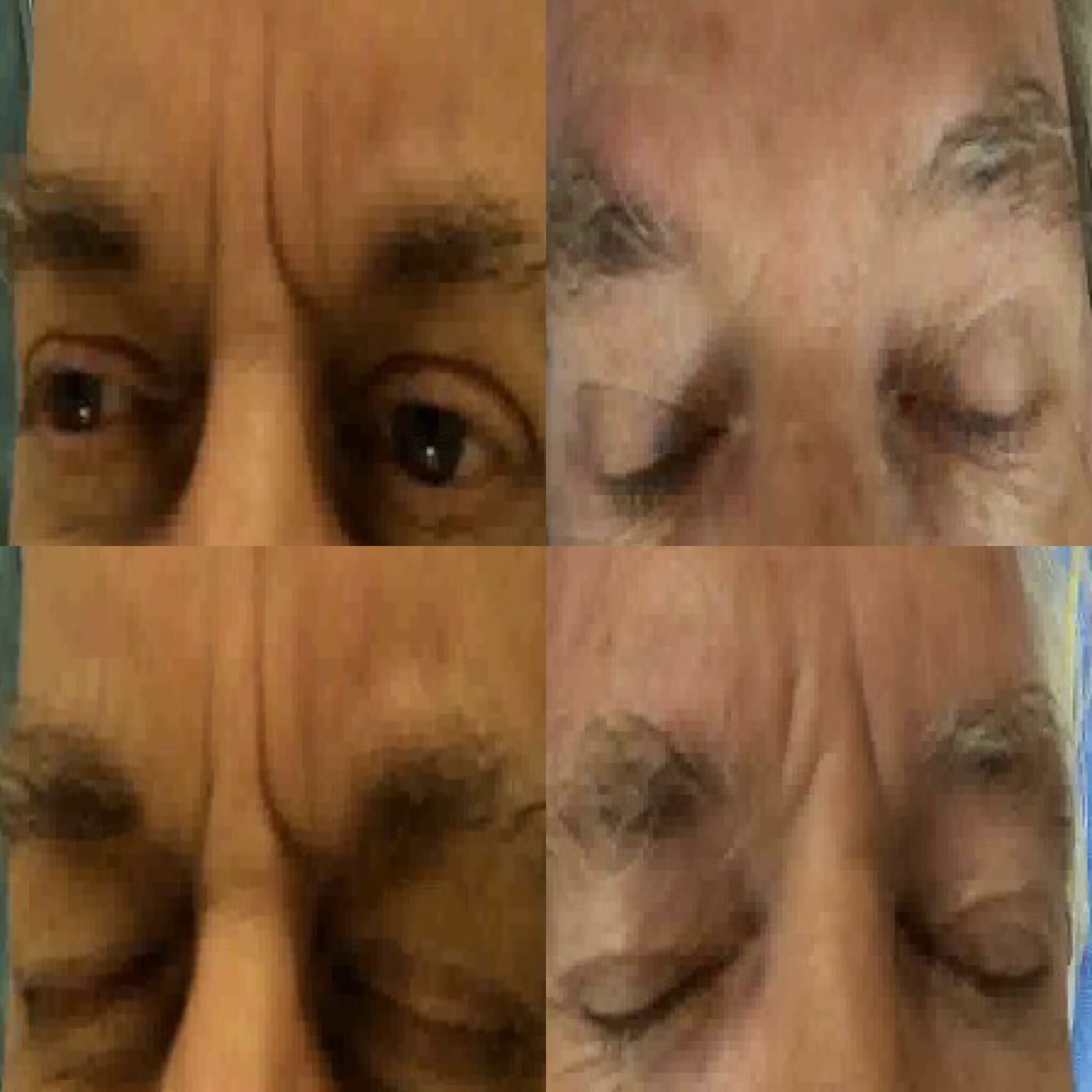}
    \caption{Sample frames from the sedation dataset after the preprcoessing steps: the  two top frames belong to the no-pain category and the bottom two frames to the pain category.}
    \label{fig:sampledata}
    \end{minipage}
\end{figure}

After obtaining the landmarks, the eyes were located using their corresponding landmarks, and the intercanthal distance was used to crop the image relative to the location of the eyes. The result cropped images were of different sizes, influence by the face-to-camera distance. The crops were resized to the same resolution of 224$\times$224 pixels. Examples of image crops from the sedation dataset can be seen in Figure~\ref{fig:sampledata}.

\addtolength{\textheight}{-3cm}   

\section{METHOD}

\subsection{Pain Detection Model}

A  ResNeXt-50-32x4d model pretrained on ImageNet was used for binary pain detection. The last fully-connected layer of ResNeXt-50-32x4d was replaced with two fully-connected layers with a ReLU activation in-between. The new layers yielded a binary output for the model. 

Three groups of models were trained. The first group were trained using the sedation dataset only; the second group were trained with the UNBC-McMaster dataset only; and the third group were trained using both datasets.

As a baseline, we re-implemented the model used by Ashraf \emph{et al.}~\cite{occludedFacesPaper}. In addition to reimplementing their model exactly (using HOG features and linear SVM), we also replaced the SVM with a Random Forest binary classifier, which yielded slightly better results on the sedation dataset. 

\subsection{Evaluation}

Leave-one-person-out cross-validation was used in all cases; except when the training set did not include images from the sedation dataset and consisted entirely of the UNBC-McMaster images. Area under the receiver operating characteristic curve (\textit{AUROC}) and under the precision-recall curve (\textit{average precision}, or \textit{AP}) were used to evaluate binary pain detection performance on the sedation dataset.

\subsection{Implementation Details}

In each cross-validation fold, The new fully-connected layers of the model were trained for 18 epochs while the other layers of the model were frozen. After 18 epochs, the last block of the model was also unfrozen for two additional epochs. The model was thus trained for a total of 20 epochs, using the Adam optimizer and with a batch size of 128. A starting learning rate of 0.005 was used, which was decayed during training by a factor of 0.1 every 10 epochs. The dataset was augmented with random horizontal flips, random affine transformation up to 30 degrees, random color jitters, and random rotations in the -10 to 10 degrees range.

When training the models with the UNBC-McMaster dataset (either with or without the sedation dataset), we binarized UNBC-McMaster PSPI scores into pain / no-pain labels using the thresholds suggested by Ashraf et al.~\cite{occludedFacesPaper}. Specifically, frames with PSPI score of 0 were considered as no-pain, frames with a PSPI score of 4 or higher were considered as pain, and frames with a PSPI score in the 1-3 range were not included in the training set. The test set is always from the sedation dataset and is not influenced by this thresholding.

\section{RESULTS AND DISCUSSION}

\begin{figure*}
\centering
\captionsetup{width=1\linewidth}
    \begin{minipage}{5.8cm}
    \includegraphics[width=1.0\textwidth]{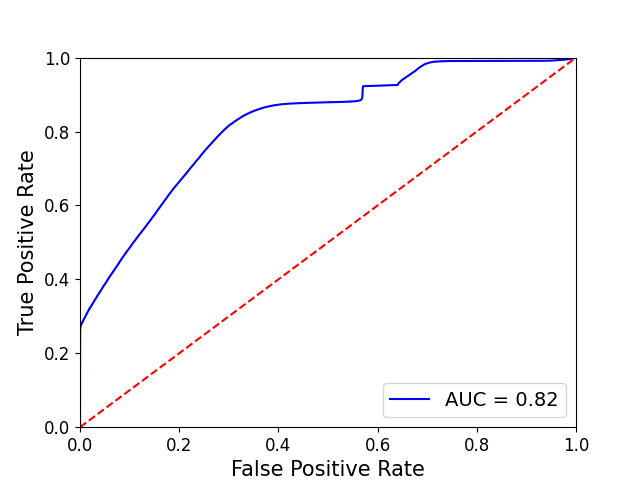}
    \centering
    \subfloat{(a)}
    \label{fig:model1ROC}
    \end{minipage}
    \begin{minipage}{5.8cm}
    \includegraphics[width=1.0\textwidth]{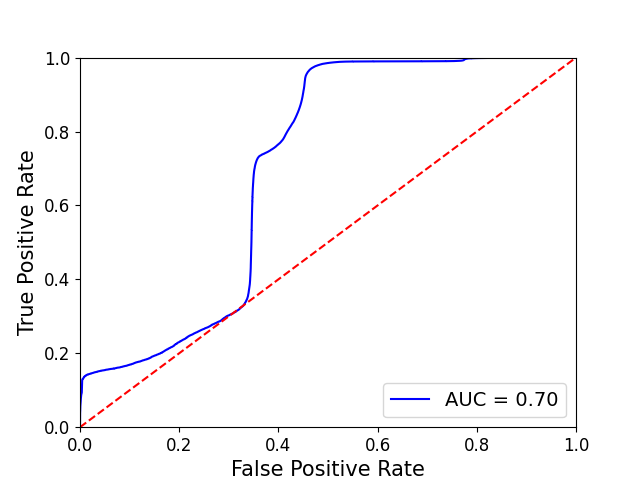}
    \centering
    \subfloat{(b)}
    \label{fig:model2ROC}
    \end{minipage}
    \begin{minipage}{5.8cm}
    \includegraphics[width=1.0\textwidth]{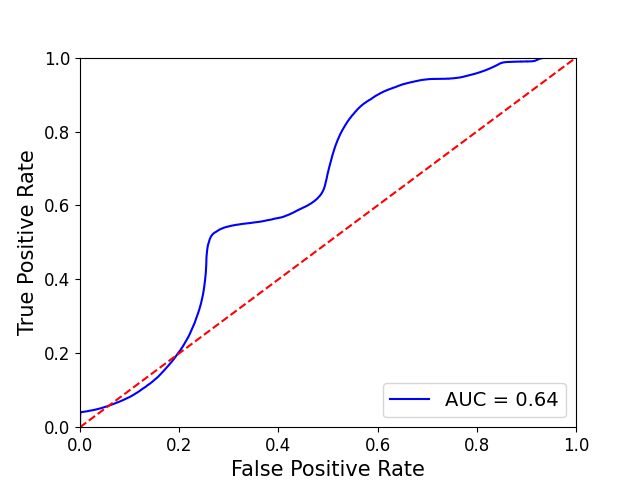}
    \centering
    \subfloat{(c)}
    \label{fig:model3ROC}
    \end{minipage}
\caption{ROC curce of models fine-tuned on a) sedatation, b) sedation and UNBC-McMaster, and c) UNBC-McMaster data.}
\label{fig:ROC}
\end{figure*}

Table~\ref{table:summaryResults} compares the per-frame results of the fine-tuned ResNeXt models, as well as baseline models of Ashraf et al.~\cite{occludedFacesPaper}, in terms of the area under the ROC curve (AUC) and average precision (AP). 

\begin{table}[h]
\begin{center}
\begin{tabular}{|c|c|c|c|c|}
\hline
Model & AUC & AP \\
\hline
ResNext Fine-tuned on Sedation & 0.75 & 0.58\\
ResNeXt Fine-tuned on Sedation \& UNBC-McMaster & 0.73 & 0.51\\
ResNeXt Fine-tuned on UNBC-McMaster & 0.65 & 0.45\\
HOG + SVM Trained on Sedation & 0.57 & 0.49 \\
HOG + Random Forest Trained on Sedation & 0.58 & 0.51 \\
\hline
\end{tabular}
\end{center}
\caption{Test performance on the sedation dataset (without temporal smoothing).}
\label{table:summaryResults}
\end{table}

Facial expressions of pain have significantly lower rate of change than the video frame rate of 30~or~60~fps. Even microexpression have an average length of an order of magnitude larger than the frame-to-frame temporal distance of $33.\overline{3}$~or~$16.\overline{6}$~ms~\cite{yan2013fast}. It is, therefore, possible to smooth per-frame prediction scores to reduce jitter and improve accuracy. Table~\ref{table:summaryResultsSmoothed} presents AUC and AP results after the prediction scores for each model were smoothed with a causal uniform  filter over the previous 30 frames for 30 fps videos, or over the previous 60 frames for 60 fps videos. We note that using the causal filter is necessary and a non-causal filter (\textit{e.g.}, uniform or Gaussian smoothing) would not be applicable in real-life dosage titration applications. We also note that the causal filter can lead to slight (under 1~second) delays in detecting expressions of pain. But such delays are inconsequential when the average length of a procedure is around 20-minutes; although they could negatively impact AUC and AP values reported in this paper.

\begin{table}[h]
\begin{center}
\begin{tabular}{|c|c|c|c|c|}
\hline
Model & AUC & AP \\
\hline
ResNext Fine-tuned on Sedation & 0.82 & 0.72\\
ResNeXt Fine-tuned on Sedation \& UNBC-McMaster & 0.70 & 0.48\\
ResNeXt Fine-tuned on UNBC-McMaster & 0.64 & 0.39\\
HOG + SVM Trained on Sedation & 0.56 & 0.43 \\
HOG + Random Forest Trained on Sedation & 0.55 & 0.44 \\
\hline
\end{tabular}
\end{center}
\caption{Test performance on the sedation dataset, when per-frame predictions are smoothed with a causal uniform filter.}
\label{table:summaryResultsSmoothed}
\end{table}


\begin{table*}[tb]
\begin{center}
\begin{tabular}{|c|c|c|c|c|c|}
\hline
 &\multicolumn{2}{|c|}{Trained only on sedation}& \multicolumn{2}{|c|}{Trained on combined datasets} & \\ \cline{2-5}
Participant & \ \ \ AUC \ \ \ & AP & \ \ \ \  AUC \ \ \ \ & AP & Number of frames (\% Pain) \\
\hline
P003  & 1.00 & 1.00 & 1.00 & 1.00 & \ 30,562 (44.11)\\
P004  & 0.99 & 0.97 & 0.96 & 0.87 & \ 92,605 \ (6.72)\\
P009  & 0.50 & 0.02 & 0.99 & 0.90 & \ 24,159 \ (1.81)\\
P011  & 0.96 & 0.94 & 1.00 & 1.00 & \ 21,998 (38.21)\\
P017  & 0.28 & 0.04 & 0.92 & 0.65 & \ 22,087 \ (5.85)\\
P018  & 0.99 & 0.99 & 0.99 & 0.99 & \ 16,668 (40.17)\\
P022  & 0.85 & 0.09 & 0.73 & 0.03 & \ 11,669 \ (0.75)\\
P029  & 1.00 & 1.00 & 1.00 & 1.00 & \ 27,707 (11.12)\\
P030  & 0.99 & 0.96 & 1.00 & 0.99 & \ 20,973 \ (5.97)\\
P033  & --   & --   & --   & --   & \ \ 9,822 \ (0.00)\\
P034  & 1.00 & 1.00 & 1.00 & 1.00 & \ 38,127 (48.58)\\
P035  & 0.99 & 1.00 & 0.77 & 0.82 & \ 70,154 (56.56)\\
P036  & --   & --   & --   & --   & \ 39,129 \ (0.00)\\
P037  & 1.00 & 1.00 & 1.00 & 1.00 & \ 57,676 (77.74)\\\hline
Total & 0.82 & 0.72 & 0.70 & 0.48 & 483,336 (29.79)\\
\hline
\end{tabular}
\end{center}
\caption{Per-participant test results for the model fine-tuned on the sedation dataset and the model trained on the combined sedation and UNBC-McMaster datasets, after the application a of causal uniform filter.}
\label{table:participantSmoothedResults}
\end{table*}

Figure~\ref{fig:ROC} shows the ROC curves for the pretrained ResNexT models fine-tuned on the sedatiaon dataset, on the UNBC-McMaster dataset, and the combined datasets, after the predictions were smoothed with the causal smoothing filter. 

Looking at the results in Tables~\ref{table:summaryResults}~and~\ref{table:summaryResultsSmoothed}, the fine-tuned deep learning models clearly outperform the baselines using HOG features. Among these, the ResNeXt model fine-tuned only on the sedation dataset has an area under the ROC curve (AUC) of 0.82 and an average precision (AP) of 0.72 (Table~\ref{table:summaryResultsSmoothed}), and outperforms the ResNeXt model fine-tuned on both datasets 
or fine-tuned only on the UNBC-McMaster dataset  
by a wide margin.

The large performance gap between the deep learning models and the baselines were expected; but the wide gap in cross-dataset performance requires an explanation. In the UNBC-McMaster dataset, the frames in which the participants have their eyes closed are mostly pain frames. In fact, the binary  facial action unit (AU) \#43 (`eyes closed') is part of the PSPI formula, and forms 1/16 (or~6.25\%) of the maximum PSPI score. 
However, because of sedation,  participants in the sedation dataset have their eyes closed in the majority of video frames, regardless of the frame label. The top right image in Figure~\ref{fig:sampledata} illustrates a sample eyes-closed no-pain frame in the sedation dataset, and the bottom right image shows a sample pain frame of the same participant.
In this dataset, distinguishing pain frames relies heavily on identifying  other types of pain expression around the eyes and in the forehead, \textit{e.g.}, wrinkles or AU~\#4 (`brow lowerer'). Therefore, the model trained only on the UNBC-McMaster data can learn eye closure as an indication of pain, which in turn leads to an over-prediction of the pain label on the sedation dataset. Differences in the filming of the videos and lighting conditions can also be contributing factors to this cross-dataset performance gap.

\begin{figure*}
\centering
\captionsetup{width=1\linewidth}
    
    \centering
    {\footnotesize P003}\par\medskip
    \includegraphics[width=.9\textwidth,height=3.7cm]{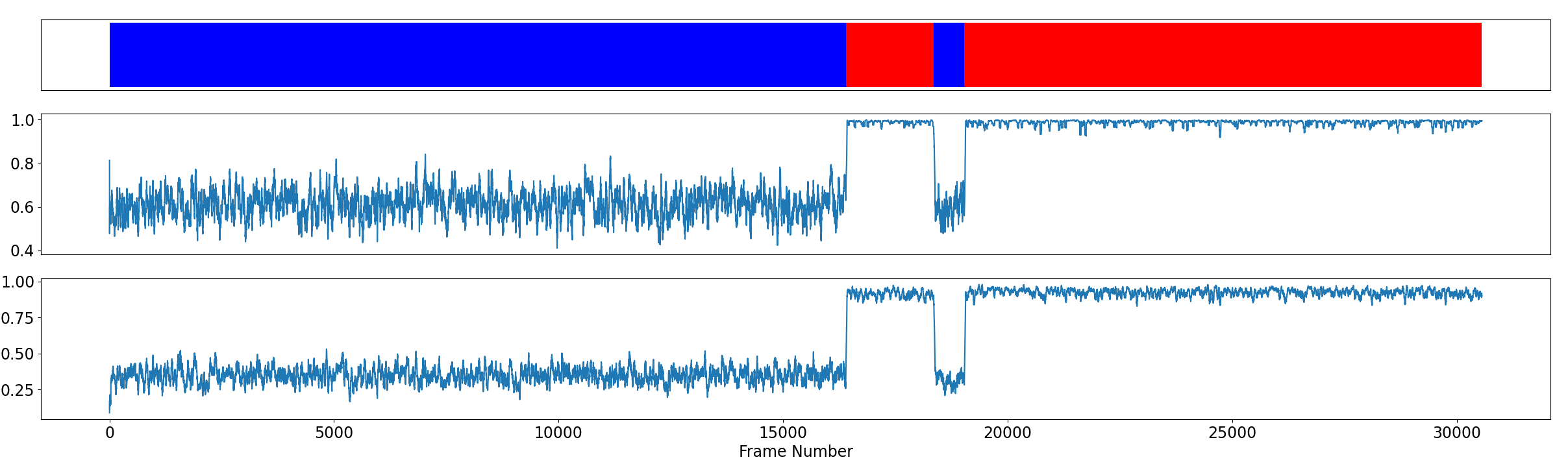}
    \centering

    \vspace{.4cm}
    
    \centering
    {\footnotesize P018}\par\medskip
    \includegraphics[width=.9\textwidth,height=3.7cm]{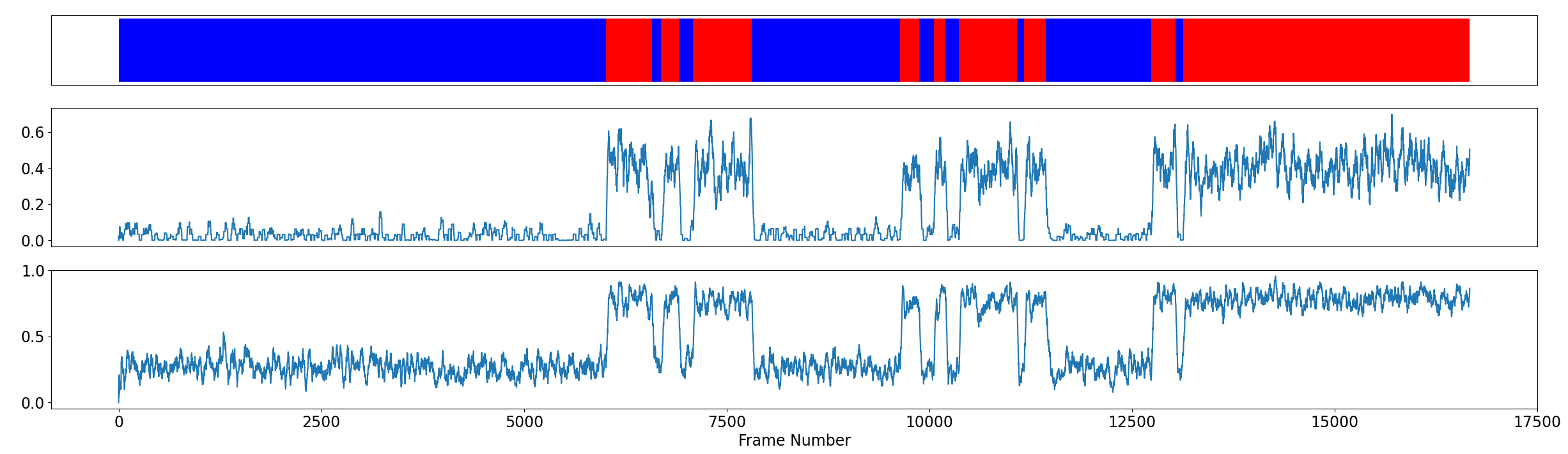}
    \centering

    \vspace{.4cm}
    
    \centering
    {\footnotesize P022}\par\medskip
    \includegraphics[width=.9\textwidth,height=3.7cm]{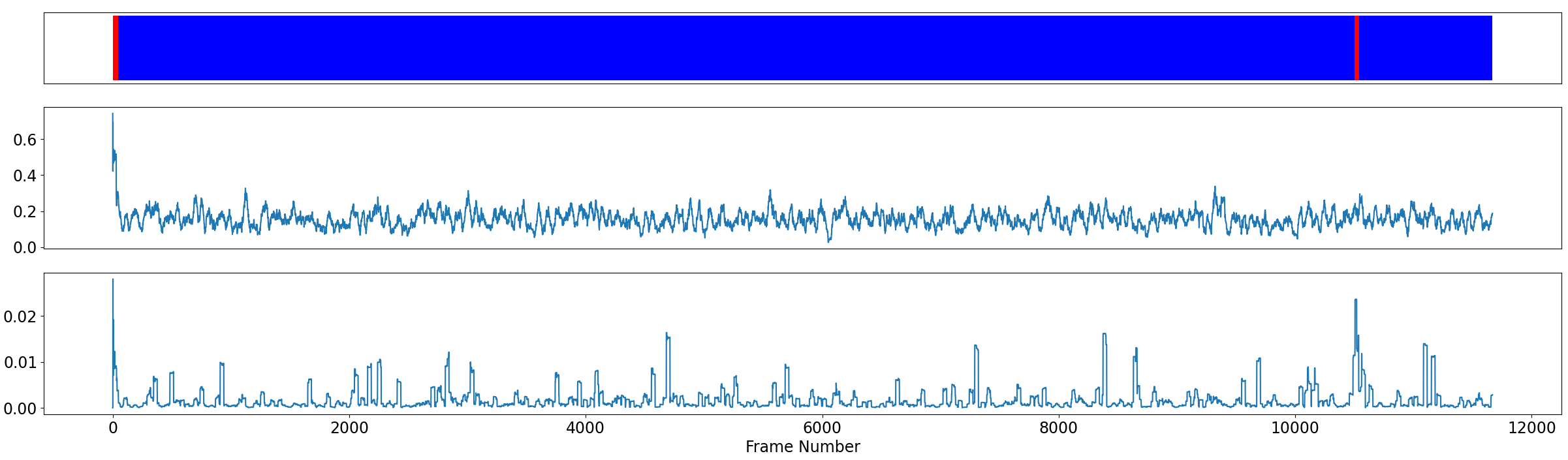}
    \centering

    \vspace{.4cm}
    
    \centering
    {\footnotesize P034}\par\medskip
    \includegraphics[width=.9\textwidth,height=3.7cm]{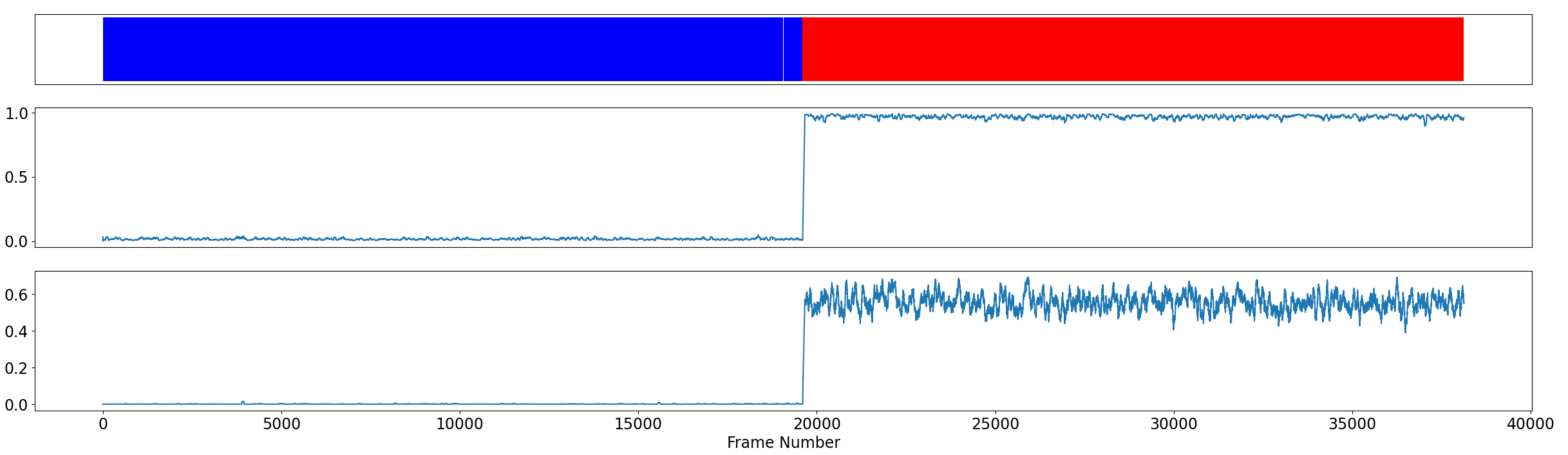}
    \centering

    \vspace{.4cm}
    
    
    \centering
    {\footnotesize P037}\par\medskip
    \includegraphics[width=.9\textwidth,height=3.7cm]{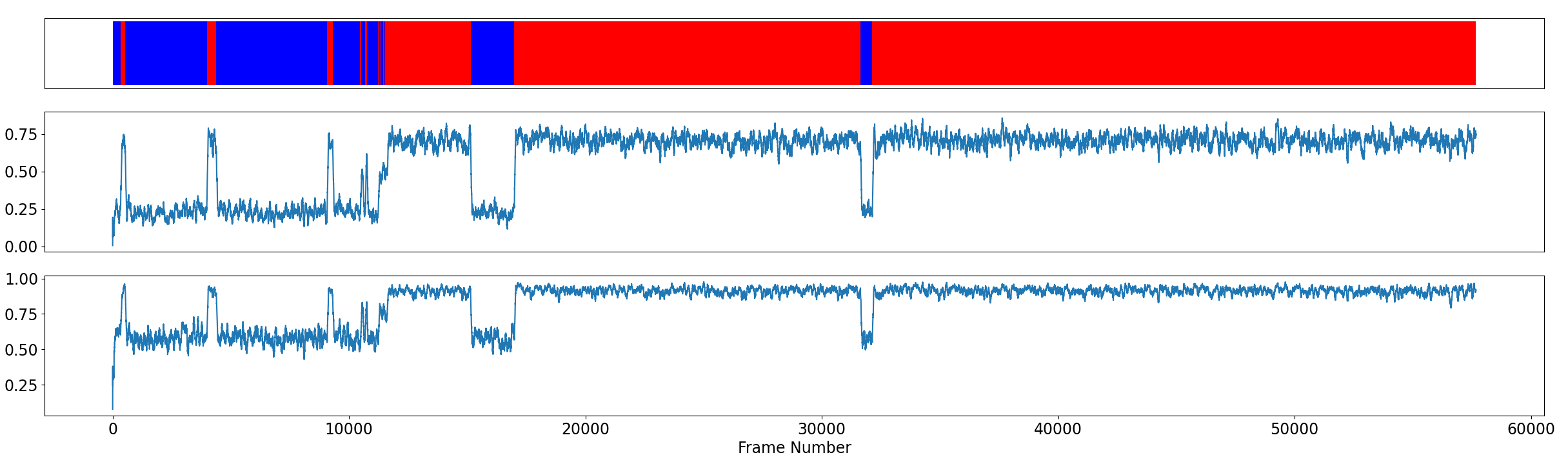}
    \centering

\caption{True pain/no-pain labels and smoothed model outputs over time for five sample participants. Each sub-figure consists of three rows: top) ground truth labels (no-pain labels in blue and pain labels in red); middle) smoothed outputs of model trained on the sedation data bottom) smoothed outputs of model trained on the sedation and UNBC-McMaster data.
}
\label{fig:temporalComparison}
\end{figure*}

Tables~\ref{table:summaryResults}~and~\ref{table:summaryResultsSmoothed} report the performance metrics (AUC and AP) when test predictions on all 14 participants
(obtained from 14 different cross-validation loops) were combined. 
Table~\ref{table:participantSmoothedResults} reports the performance in each participant, for the ResNeXt models fine-tuned on the sedation data alone, and the one fine-tuned on the combined sedation + UNBC-McMaster data.
Two (2) of the participants (P033 and P036) did not show any expressions of pain during the medical procedure and, as such, do not have associated AUC and AP values. It should be noted that test results on data from both participants still contributed to the total AUC and AP values reported in Tables~\ref{table:summaryResults}~and~\ref{table:summaryResultsSmoothed}. Of the remaining (14-2=)12 participants, perfect or near perfect pain detection was achieved on nine (9) participants, using the model fine-tuned only on the sedation data. Performance was poor on the three (3) remaining participants, namely P009, P017, and P022. Looking at the rightmost column of Table~\ref{table:participantSmoothedResults}, it can be observed that, apart from the two participants with zero pain frames, the percentage of pain frames is the lowest among these three participants. 
Figure~\ref{fig:temporalComparison} illustrates ground truth (pain/no-pain) labels and smoothed model outputs over time for five sample participants, including one for whom the model obtained poor pain detection results (P022). 


\section{CONCLUSIONS AND FUTURE WORKS}


In this paper, we explored computer vision approaches for detecting pain in partially occluded faces of sedated people. Fine-tuning a deep learning model outperformed the baseline when tested on a video dataset collected from masked faces of patients undergoing a medical procedure with sedation, and showed the viability of deep learning methods in automatic pain detection for the target population. We also showed that fine-tuning the model with a different dataset did not obtain good results on the sedation videos and qualitatively explained the reason why. 
Video data from more participants is currently being collected to increase variability of the dataset and improve the generalization of pain detection models. Future work with a larger dataset will examine performance in different subgroups to investigate sensitivity to skin-tone and gender.

\section{ACKNOWLEDGMENTS}

The authors would like to thank the AGE-WELL Network of Centres of Excellence, Natural Sciences and Engineering Research Council of Canada (NSERC), and KITE Research Institute, Toronto Rehabilitation Institute – University Health Network.


{\small
\bibliographystyle{ieee}
\bibliography{egbib}
}

\end{document}